# Feasibility of Supervised Machine Learning for Cloud Security

Deval Bhamare
Qatar University,
Doha, Qatar
devalb@qu.edu.qa

Tara Salman
Washington University,
St. Louis, USA
tara.salman@wustl.edu

Mohammed Samaka
Qatar University,
Doha, Qatar
samaka.m@qu.edu.qa

Aiman Erbad
Qatar University,
Doha, Qatar
aerbad@qu.edu.qa

Raj Jain
Washington University,
St. Louis, USA
jain@wustl.edu

*Abstract: Cloud computing is gaining significant attention, however, security is the biggest hurdle in its wide acceptance. Users of cloud services are under constant fear of data loss, security threats and availability issues. Recently, learning-based methods for security applications are gaining popularity in the literature with the advents in machine learning techniques. However, the major challenge in these methods is obtaining real-time and unbiased datasets. Many datasets are internal and cannot be shared due to privacy issues or may lack certain statistical characteristics. As a result of this, researchers prefer to generate datasets for training and testing purpose in the simulated or closed experimental environments which may lack comprehensiveness. Machine learning models trained with such a single dataset generally result in a semantic gap between results and their application. There is a dearth of research work which demonstrates the effectiveness of these models across multiple datasets obtained in different environments. We argue that it is necessary to test the robustness of the machine learning models, especially in diversified operating conditions, which are prevalent in cloud scenarios. In this work, we use the UNSW dataset to train the supervised machine learning models. We then test these models with ISOT dataset. We present our results and argue that more research in the field of machine learning is still required for its applicability to the cloud security.*

## I. Introduction

Recently cloud computing is gaining significant traction and virtualized data centers are becoming popular as a cost-effective infrastructure and solution for enterprise applications. Infrastructure as a Service (IaaS), Platform as a Service (PaaS) and Software as a Service (SaaS) are being widely deployed and utilized by the end users. In this way, users neither require knowledge, control, and ownership in the computing infrastructure nor they need to host, control or own an infrastructure in order to deploy their applications. Instead, they simply access or rent the hardware or software paying only for what they use. The possibility of paying-as-you-go along with on-demand elastic operations by cloud hosting providers is gaining popularity in enterprise computing model [1]. Regardless of its advantages, the transition to this computing paradigm is hampered by major security issues, which are the subject of many recent studies. Recently there has been much interest in Machine Learning (ML) techniques for network and cloud security. The fact has been demonstrated in the surveys, including one by Tsai et al. [2] and Fernandes et al. [3]. These works demonstrate the popularity of the machine learning techniques among the scholars in the context of cloud and network security.

The general approach which can be observed towards developing ML based security applications is, training the models using labeled network traces obtained with some experimental environment, with a comprehensive set of intrusions and abnormal behavior along with the normal behavior [4]. Generally different sets of features are extracted from these traces and are used to train the models. A detailed description of such features is out of scope of this article and readers are advised to refer to works in [8-10] for more details. Another portion of the same dataset is then used to test these trained models [25, 34].

This approach itself is a significant challenge and has some flaws. Security related datasets are extremely rare, mostly due to privacy issues. Hence, machine learning models are trained and evaluated over experimentally generated datasets that lacks sufficient comprehensiveness [4, 30]. As a result of this, models performing well with one particular dataset may fail to perform with other datasets. This is due to the fact that user packets need to travel through different data centers/clouds, distributed across multiple locations operating in diversified environments, especially, with the advents of the NFV and SFC networking paradigms [36]. Furthermore, most learning datasets include only specific types of attacks while neglecting others [4, 7]. The authors in [4] argue that the above approach severely impairs the evaluation of machine learning models, particularly affecting anomaly-based detectors. The authors also claim that, "despite the significant contributions of DARPA [5] and KDD [6] datasets in the intrusion detection domain, their accuracy and ability to reflect real-world conditions has been extensively criticized in [7]."

There is a dearth of research works which focus on training and testing of the machine learning models over different datasets. This is in fact necessary to test the robustness and applicability of the machine learning algorithms in real-time scenarios. Addressing this particular flaw in recent studies, in this work, we analyze the performance of major supervised machine learning algorithms with two different datasets, namely, UNSW [8, 9] and ISOT [10]. Both the datasets have been obtained in simulated cloud environments. This variety of traffic with two completely different datasets serves as a good example of day-to-day use of enterprise cloud networks [8, 10]. To be precise, we compare regression, decision trees, Naïve Bayes, and Support Vector Machines (SVM) techniques. We chose these algorithms, as they are widely used in different fields, including network and cloud security [12-15, 34].

First, the models are trained using UNSW training dataset. The trained models are then tested using UNSW test and ISOT datasets. It is to be noted that UNSW training and test datasets are obtained using same experimental environment, however, ISOT dataset has been obtained using completely different experimental setup (please refer to Section IV for the detailed description of the datasets). We argue considering multiple datasets will provide the required comprehensiveness in testing ML models which has been missing so far in the research works [4, 7]. Such testing of the learned models can provide a sense of robustness and applicability of the learned models in real scenarios. We present our results to demonstrate the need for further research in the field of supervised machine learning and its applicability to cloud and network security. The rest of the paper has been organized as follows. In Section II we discuss the state of the art. Section III provides a brief description for the datasets under consideration. In Section IV we present our findings. Section V finally concludes the paper.

## II. Related Work

Cloud security has been studied for a long time in the literature. A detailed analysis of the security threats in the cloud computing environment is given in [12, 13]. Recently, there has been a trend to apply ML techniques for network and cloud security as well [14, 15]. For example, SVM based approach has been proposed in [16, 17]. Artificial Neural Network (ANN) based approach to build the security model has been proposed in [18, 19]. Multilayer perceptron (MLP), a widely used model has been proposed in [19]. A decision tree based approach has been proved to be efficient to detect network anomalies [20, 21]. Furthermore, hybrid of two more ML models has been studied in the literature as well. Results demonstrate the superiority of the hybrid approaches [22].

A hybrid approach of decision trees and SVM has been proposed in [23]. The authors in [24] propose Bayesian network based model to detect the network threats. Data mining approaches have been proved efficient in [25] for anomaly detection in networks. On the other hand, recent advances in the computing technologies have aided attackers as well. For example, evolution of Denial of Service (DoS) attacks to Distributed DoS (DDoS) attacks [26, 27]. Cloud Security Alliance has identified DDoS attack as one of the nine major threats [28]. A detailed survey of other possible threats in cloud environment and intrusion detection techniques is given in [29].

A common blunder in the anomaly detection works discussed so far is the assumption of same operational environments. The developed models are trained and tested on the datasets obtained within the same experimental setup. Generally, packet dumps are captured and some sets of features are extracted [25, 34]. The values of these features depend on the experimental environment. However, this may not be the case in real-time scenarios. Simulated environments within the labs are not able to cover all the possible scenarios. Moreover, as network behaviors and patterns change, intrusions evolve, and new attack types come to evidence. Hence, it has become necessary to shift to dynamic learning models. In other words, we should consider moving from static, sub-optimal and single datasets towards more dynamically generated hybrid datasets to reflect the real-time traffic conditions. Same fact has been confirmed by [4-7].

Hence, we argue that machine learning models trained with particular datasets need to be tested with completely different datasets to test their robustness. In this work, we train different supervised machine learning models using the labeled UNSW datasets. We then test these models with ISOT dataset obtained from completely different experimental setups and environments. In the next section, we quickly glance through the datasets used in this work.

## III. Datasets Description

The UNSW-NB-15 dataset [8] was created using an IXIA PerfectStorm tool in the Cyber Range Lab of the Australian Centre for Cyber Security (ACCS) to generate a hybrid of the realistic modern normal activities and the synthetic contemporary attack behaviors from network traffic. A *tcpdump* tool was used to capture 100 GB of a raw network traffic. This dataset has nine types of attacks, namely, Fuzzers, Analysis, Backdoors, DoS, Exploits, Generic, Reconnaissance, Shellcode and Worms. Argus, Bro-IDS network monitoring tools are used and twelve algorithms are developed to generate totally 49 features with the class label [9]. For more detailed description of the features and attack-types, readers are advised to read Tables 1 to 5 (Section 2) in [8, 9].

Parts of this dataset are configured as training set and testing set, namely, "*UNSW_NB15_training-set.csv*" and "*UNSW_NB15_testing-set.csv*," respectively, which has been used in this work to develop the machine learning models. The number of records in the training set is 175,341 records and the testing set is 82,332 records. Table I below gives the statistics for the anomalous and normal packets in these two datasets.

Table I: UNSW dataset statistics

| Dataset | Total Records | Normal | Anomalous |
|---|---|---|---|
| Training | 175342 | 56000 | 119341 |
| Testing | 82332 | 37000 | 45332 |
| Total | 257674 | 93000 | 164674 |
| Percentages | 100% | 36.1% | 63.9% |

ISOT dataset [10] has been obtained from two separate datasets containing malicious traffic from the French chapter of the *Honeynet* project involving the Storm and *Waledac* botnets, respectively [31]. In additions, to represent non-malicious, everyday usage traffic, two different datasets, one from the Traffic Lab at Ericsson Research in Hungary [32] and the other from the Lawrence Berkeley National Lab (LBNL) [33] have been incorporated. The data have been recorded over a three month period, from October 2004 to January 2005 covering 22 subnets. Table II below summarizes the statistics for ISOT dataset. In the next section we present our results obtained using these two datasets with some major supervised machine learning algorithms.

Table II: ISOT dataset statistics

| Traffic Type | Unique Flows | Percentages |
|---|---|---|
| Malicious | 55904 | 3.33% |
| Normal | 1619520 | 96.66% |
| Total | 1675424 | 100% |

## IV. Results and Analysis

In this section, initially we compare performance of the supervised machine learning techniques using UNSW dataset. Later on we test the robustness of these learned models with ISOT dataset. As mentioned earlier, we have chosen a specific algorithm from major supervised machine learning schemes. For example, we have chosen J48 from decision trees [23], Naïve Bayes from Bayesian networks [24], logistic regression from regression techniques (LR) [35] and SVM with three different kernels, which are SVM-RBF, SVM-Polynomial, SVM-Linear [16, 23]. We chose these algorithms, as they represent major supervised schemes and are widely used in different areas including network security [34]. Initially we train aforementioned supervised machine learning algorithms using UNSW training dataset and WEKA tool [11]. Then we compare the performance the algorithms over UNSW testing dataset. The results are presented below.

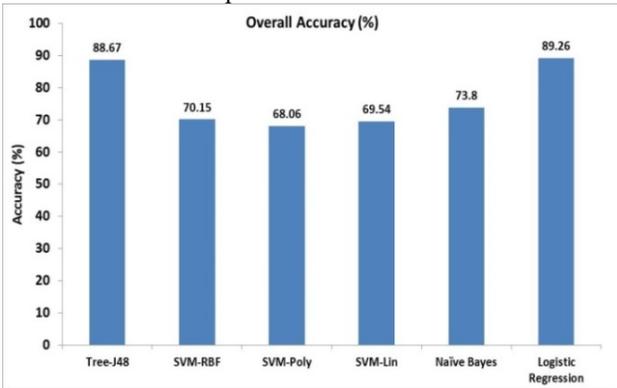

Fig. 1: Overall accuracy with UNSW Dataset

Fig. 1 displays the overall performance of the machine learning techniques mentioned earlier. We observe that with the UNSW training and testing datasets, logistic regression performs the best with 89.26% accuracy, followed by J48 algorithm, with 88.67% accuracy. The polynomial SVM algorithm has the least accuracy with 68.06% accuracy. However, overall accuracy is not enough to compare the performance of these algorithms. It is imperative to observe true positive, false positive, true negative and false negative rates as well. Table III below explains the meaning of these terms in brief. Here zero represents a normal and one represents an anomalous packet.

Table III: Prediction conditions in machine learning

| Total Population | Predicted Condition Positive (Anomalous) | Predicted Condition Negative (Normal) |
|---|---|---|
| Positive (1, Anomalous) | True Positive (TP) | False Negative (FN) |
| Negative (0, Normal) | False Positive (FP) | True Negative (TN) |

Fig. 2 below shows the TP and FN percentages for these algorithms. We observe that logistic regression performs the best, with a 93.7 % rate for identifying anomalous packet correctly (TP), while it identifies 6.3% of anomalous packets as normal (FN), which is an error case. J48 follows with the TP and FN percentages as 84.7 and 15.3 respectively. Naïve Bayes performs the worst with the percentages as 56.2 and 43.8 respectively.

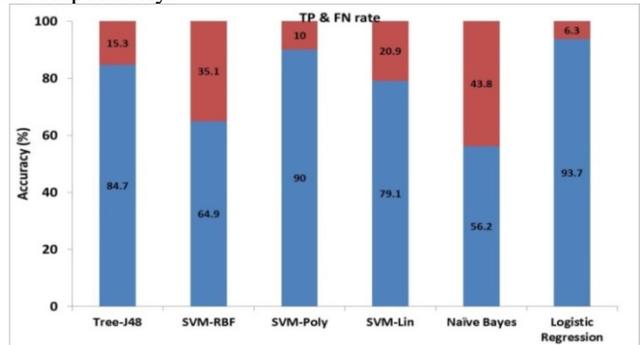

Fig. 2: TP and FN rate with UNSW dataset

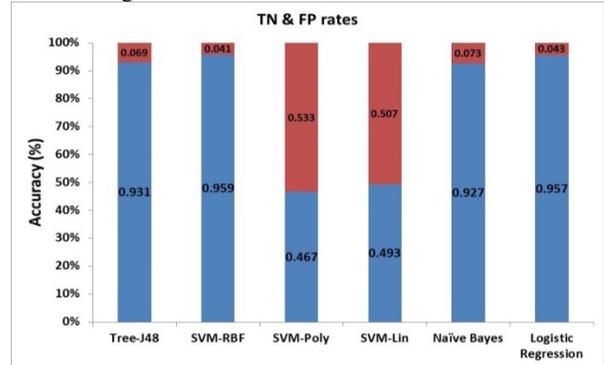

Fig. 3: TN and FP rate UNSW dataset

Graphs for TN and FP are displayed in Fig. 3. We observe that SVM-RBF performs best with 95.9% accuracy in identifying the normal packets correctly (TN) followed by linear regression with 95.7%. Since LR performs the best overall, we investigate further with LR to improve its performance in identifying the anomalous packets correctly. In other words, we aim to increase TP rate for LR, since identifying anomalous packets has the highest importance in security applications.

We extract the parameters for linear regression from the model generated by WEKA and implement it separately using

OCTAVE. We then vary the probability threshold to categorize packet as normal versus anomalous packet. By default, the value is 0.5, that is, if the final probability for the packet being normal is greater than 0.5, then it is classified as normal, else it is identified as anomalous. However, in cloud security, the major concern is identifying as many anomalous packets as possible because the penalty for false negative in security is very high. Hence we vary the threshold frequency, from 0.1 to 0.9 and plot the graphs for TP and TN. As expected, in Fig. 3, with the increase in the threshold, TP rate increases, that is percentage for identification of anomalous packets increases while that of TN decreases. At threshold value 0.7 and 0.8, we observe that TP percentage is around 97%, which is an accuracy of detecting anomalous packets is as high as 97%. Also, accuracy in identifying the normal packets (TN) is around 80%, which is acceptable. Hence we conclude that the probability threshold can be kept higher than 0.5, which is around 0.7 to 0.8 to detect anomalous packets more accurately.

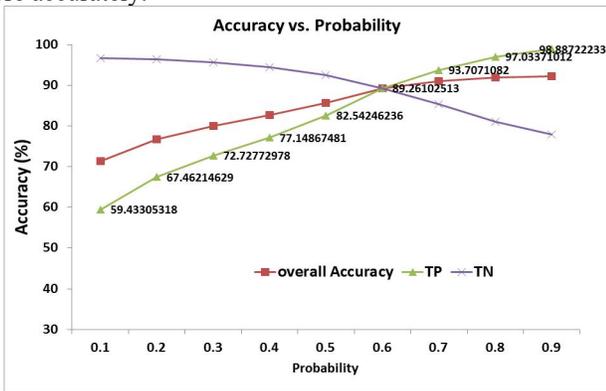

Fig. 4: Varying Frequency threshold with UNSW dataset

We have obtained satisfactory results with UNSW training and testing datasets, however, as mentioned earlier, we propose to test the robustness of the models with another dataset obtained in a completely different environment. For our results, we use ISOT dataset. We chose three techniques, which performed best with UNSW dataset that is J48, SVM-RBF and LR. Fig. 5 shows the overall accuracy of these models with the ISOT dataset. As we can see, the overall accuracy is still impressive, that is around 95% with J48 and LR and around 90% with SVM. In addition, we performed similar experiments with LR by varying the threshold to increase TP rate and obtained results as shown in Fig. 7. As we observe, even at the highest value of threshold at 0.9, the TP rate (identifying the anomalous packets) is not more than 54%, which is not acceptable. We recall that with the UNSW dataset, the value was as high as 97%.

However, results displayed in Fig. 6 demonstrate that the accuracy of these models to detect the anomalous packet have deteriorated. For example, the highest accuracy for TP is around 43%, which is very low as compared against the one in Fig. 2 which is 93.7%. Such a low accuracy in identifying the anomalous packets may prove disastrous in real-time scenarios. This is a significant evidence that the models trained and tested with data from one single environment may not perform well with other datasets in the network or cloud security scenario.

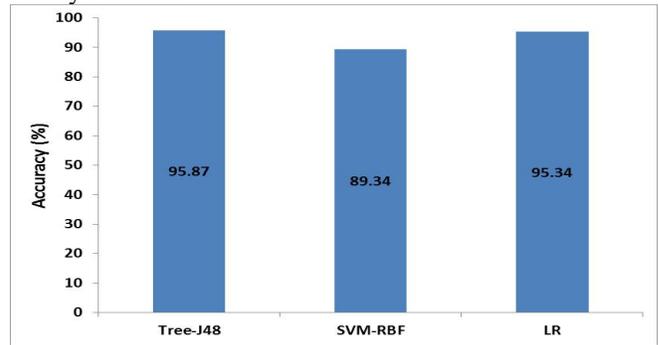

Fig. 5: Overall accuracy with ISOT Dataset

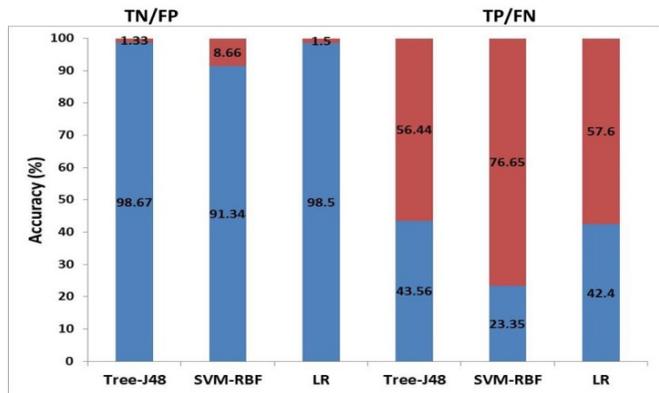

Fig. 6: TP and TN rate with ISOT dataset

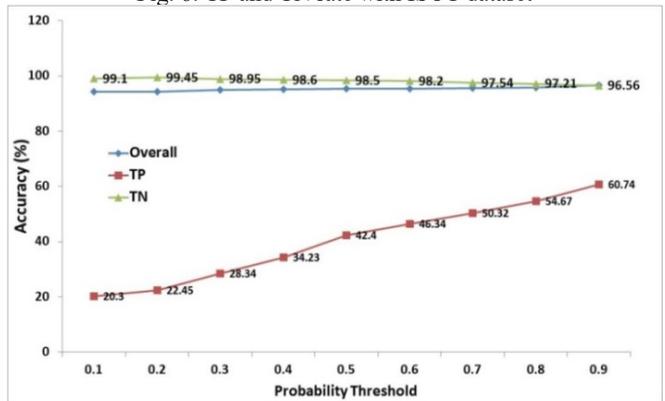

Fig. 7: Varying Frequency threshold with ISOT dataset

This may happen due to different network settings or new attack behavior that was not considered in the training dataset. Moreover, this may be a result of the fact that supervised machine learning techniques performs well with a large number of positive (1) as well as negative (0) samples in the training dataset. However, this may not be the case with anomaly detection, where anomalous events are rare and have a very small number of positive samples [34]. With the help of the results we argue that more research needs to be carried out in the domain of machine learning as far as network and cloud security is concerned. It may include exploring the unsupervised machine learning techniques or exploring clustering and other related techniques for anomaly detection.

## V. Conclusions

Our work focuses on the imbalance between the extensive amount of research on supervised ML techniques and their applicability to the real-time scenarios. With reference to the above results, we conclude that the supervised machine learning models that perform well with a particular dataset, may or may not perform satisfactory with totally different datasets generated with different simulation or experimental conditions and environments. In addition, a single dataset can't include all types of attacks. This is a prevalent condition in the cloud scenarios, especially with the advent in the new networking paradigms such as Network Function Virtualization (NFV) and Service Function Chaining (SFC), where packets need to travel more than one datacenter and/or clouds, operating in different conditions. Hence we conclude that, supervised ML techniques need significant rework to perform better in the context of cloud security. More research is needed in the areas such as anomaly detection, clustering and/or unsupervised machine learning in the future.

ACKNOWLEDGEMENT

This publication was made possible by the NPRP grant # 6 - 901 - 2 - 370 from the Qatar National Research Fund (a member of The Qatar Foundation). The statements made herein are solely the responsibility of the author [s].